\documentclass[10pt,twocolumn,letterpaper]{article}

\usepackage{cvpr}
\usepackage{times}
\usepackage{epsfig}
\usepackage{graphicx}
\usepackage{amsmath}
\usepackage{amssymb}
\usepackage{upgreek}
\usepackage{booktabs}
\usepackage{multirow}
\usepackage{multicol}
\usepackage{caption}
\usepackage{subcaption}
\usepackage{bbding}
\usepackage{stfloats}

\usepackage{color}

\usepackage[pagebackref=true,breaklinks=true,letterpaper=true,colorlinks,bookmarks=false]{hyperref}

\newcommand\blfootnote[1]{%
	\begingroup
	\renewcommand\thefootnote{}\footnote{#1}%
	\addtocounter{footnote}{-1}%
	\endgroup
}

\cvprfinalcopy 

\def\cvprPaperID{4311} 

\ifcvprfinal\fi
\begin{document}
	
	\title{Deformable Siamese Attention Networks for Visual Object Tracking}
	
	\author{Yuechen Yu, Yilei Xiong, Weilin Huang$^\dagger$, Matthew R. Scott\\
		Malong Technologies\\
		{\tt\small \{rogyu, yilxiong, whuang, mscott\}@malong.com}
	}
	
	\maketitle
	\thispagestyle{empty}
	\begin{abstract}
		
		Siamese-based trackers have achieved excellent performance on visual object tracking. However, the target template is not updated online, and the features of the target template and search image are computed independently in a Siamese architecture.
		%
		%
		In this paper, we propose Deformable Siamese Attention Networks, referred to as SiamAttn, 
		by introducing a new Siamese attention mechanism that computes deformable self-attention and cross-attention.
		The self-attention learns strong context information via spatial attention, and selectively emphasizes interdependent channel-wise features with channel attention. The cross-attention is capable of aggregating rich contextual interdependencies between the target template and the search image, providing an implicit manner to adaptively update the target template.
		%
		%
		In addition, we design a region refinement module that computes depth-wise cross correlations between the attentional features for more accurate tracking. %
		We conduct experiments on six benchmarks, where our method achieves new state-of-the-art results, outperforming the strong baseline, SiamRPN++~\cite{li2019siamrpn++}, by 0.464$\rightarrow$0.537 and 0.415$\rightarrow$0.470 EAO on VOT 2016 and 2018.	 Our code is available at: https://github.com/msight-tech/research-siamattn.
		\blfootnote{Corresponding author: whuang@malong.com}		
	\end{abstract}
	
	\section{Introduction}\label{sec:intro}
	Visual object tracking aims to track a given target object at each frame over a video sequence. It is a fundamental task in computer vision~\cite{henriques2014high,held2016learning,kalal2011tracking}, and has numerous practical applications, such as automatic driving~\cite{lee2015road}, human-computer interaction~\cite{liu2012hand}, robot sensing, \etc. Recent efforts have been devoted to improving the performance of visual object trackers. However, developing a fast, accurate and robust tracker is still highly challenging due to the vast amount of deformations, motions and occlusions that often occur on video objects with complex background
	~\cite{wu2013online,kristan2018sixth,fan2019lasot}.
	
	\begin{figure}[t]
		\begin{center}
			\includegraphics[width=1.0\linewidth]{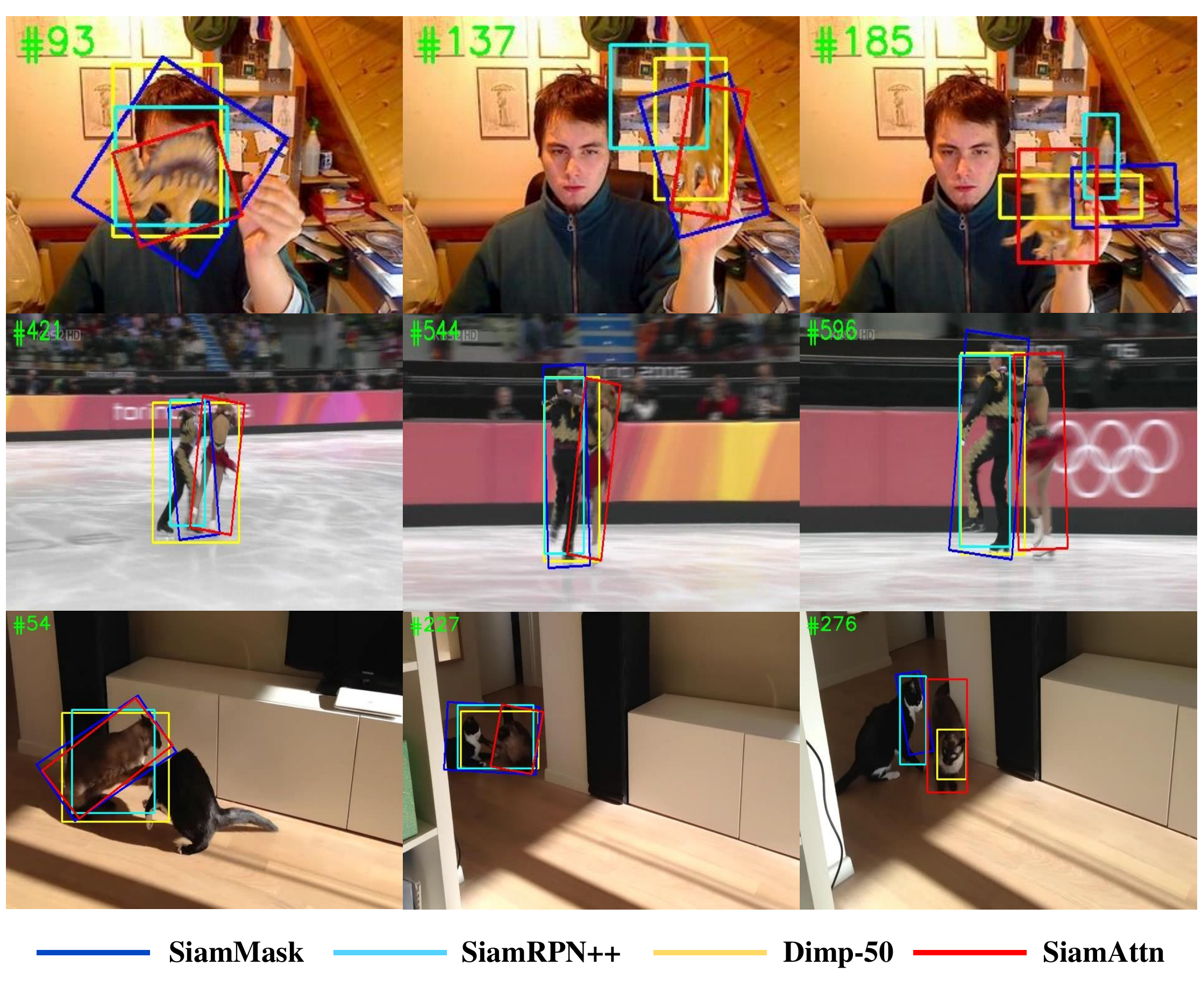}
		\end{center}
		\vspace{-5mm}
		\caption{Tracking results of our deformable Siamese attention networks (SiamAttn) with three state-of-the-art trackers. Our results are more accurate, and are robust to appearance changes, complex background and close distractors with occlusions. Fig.~\ref{fig:attention-examples} further shows the strong discriminative features learned by our Siamese attention module. }
		\label{fig:exampleshows}
		\vspace{-10pt}
	\end{figure}	
	
	Deep learning technologies have significantly advanced the task of visual object tracking, by providing the strong capacity of learning powerful deep features. For example, Bertinetto \etal~\cite{bertinetto2016fully} first introduced Siamese networks for visual tracking. 
	Since then, object trackers built on Siamese networks and object detection frameworks have achieved the state-of-the-art performance, such as SiamRPN~\cite{li2018high}, SiamRPN++~\cite{li2019siamrpn++}, and SiamMask~\cite{wang2019fast}.
	The Siamese-based trackers formulate the problem of visual object tracking as a matching problem by computing cross-correlation similarities between a target template and a search region, which transforms the tracking problem into finding the target object from an image region by computing the highest visual similarity~\cite{bertinetto2016fully,li2018high,li2019siamrpn++,wang2019fast,zhu2018distractor}. 
	Therefore, it casts the tracking problem into a Region Proposal Network (RPN)~\cite{girshick2015fast} based detection framework by leveraging Siamese networks, which is the key to boost the performance of recent deep trackers.
	
	Siamese-based trackers are trained completely offline by using massive frame pairs collected from videos, and thus the target template can not be updated online. This makes it difficult to precisely track the targets with large appearance variations, significant deformations, or occlusions, which inevitably increase the risk of tracking drift. Furthermore, the convolutional features of the target object and the search image are computed independently in Siamese architecture, where background context information is completely discarded in target features, but is of great importance to distinguish the target from close distractors and complex backgrounds. Recent work attempted to enhance the target representation by integrating the features of previous targets~\cite{Yang2018,Guo2017}, but the discriminative context information from the background is ignored. Alternatively, we introduce a new Siamese attention mechanism that encodes rich background context into the target representation by computing cross-attention in the Siamese networks.
	
	Recently, the attention mechanism was introduced to visual object tracking in~\cite{wang2018learning,zhu2018end}, which inspired the current work. However, the attentions and deep features of the target template and the search image are computed separately in~\cite{wang2018learning,zhu2018end}, which limits the potential performance of the Siamese architecture. In this work, we propose Deformable Siamese Attention Networks, referred as SiamAttn, to improve the feature learning capability of Siamese-based trackers. We present a new deformable Siamese attention which can improve the target representation with strong robustness to large appearance variations, and also enhance the target discriminability against distractors and complex backgrounds, resulting in more accurate and stable tracking, as shown in Fig.~\ref{fig:exampleshows}. The main contributions of this work are:
	
	\begin{itemize}	
		
		\vspace{-5pt}
		\item[--] We introduce a new Siamese attention mechanism that computes deformable self-attention and cross-attention jointly. The self-attention captures rich context information via spatial attention, and at the same time, selectively enhances interdependent channel-wise features with channel attention. The cross-attention aggregates meaningful contextual interdependencies between the target template and the search image, which are  encoded into the target template adaptively to improve discriminability.
		
		\vspace{-5pt}
		\item[--] We design a region refinement module by computing depth-wise cross correlations between the attentional features. This further enhances feature representations, leading to more accurate tracking by generating both bounding box and mask of the object.
		
		\vspace{-5pt}
		\item[--] Our method achieves new state-of-the-art results on six benchmarks. It outperforms recent strong baselines, such as SiamRPN++~\cite{li2019siamrpn++} and SiamMask~\cite{wang2019fast}, by a large margin. For example, it improves SiamRPN++ by 0.464$\rightarrow$0.537 and 0.415$\rightarrow$0.470 (EAO) on VOT 2016 and 2018, while keeping real-time running speed using ResNet-50~\cite{he2016deep}.
	\end{itemize}
	
	\section{Related Work}\label{sec:rel}
	
	Correlation filter based trackers have been widely used since MOSSE~\cite{bolme2010visual}, due to their efficiency and expansibility. However, the tracking object can be continuously improved online, which inevitably limits the representation ability of such trackers. 
	Deep learning technologies provide a powerful tool with the ability to learn strong deep representations,	and recent work attempted to incorporate the correlation filter framework with such features learning capability, such as 
	MDNet~\cite{nam2016learning}, C-COT~\cite{danelljan2016beyond}, ECO~\cite{danelljan2017eco} and GFS-DCF~\cite{xu2019joint}. 
	
	There is another trend to build trackers on Siamese networks, by learning from massive data offline. 
	%
	Bertinetto \etal~\cite{bertinetto2016fully} first introduced SiamFC for visual tracking, by using Siamese networks to measure the similarity between target and search image. 
	Then Li \etal~\cite{li2018high} applied a region proposal network (RPN)~\cite{girshick2015fast} into Siamese networks, referred as SiamRPN. 
	Zhu \etal~\cite{zhu2018distractor} extended the SiamRPN by developing distractor-aware training. 
	Recently, SiamDW-RPN~\cite{zhang2019deeper} and SiamRPN++ \cite{li2019siamrpn++} were proposed, which allow the Siamese-based trackers to explore deeper networks,
	while Wang \etal~\cite{wang2019fast} developed a SiamMask that incorporates instance segmentation into tracking. 
	Our work is related to that of ~\cite{fan2019siamese} where a C-RPN was developed to progressively refine the location of target with a sequence of RPNs, but we design a new module that only refines a single output region, which is particularly lightweight and can be integrated into very deep Siamese networks. 
	
	\begin{figure*}
		\begin{center}
			\includegraphics[width=0.9\linewidth]{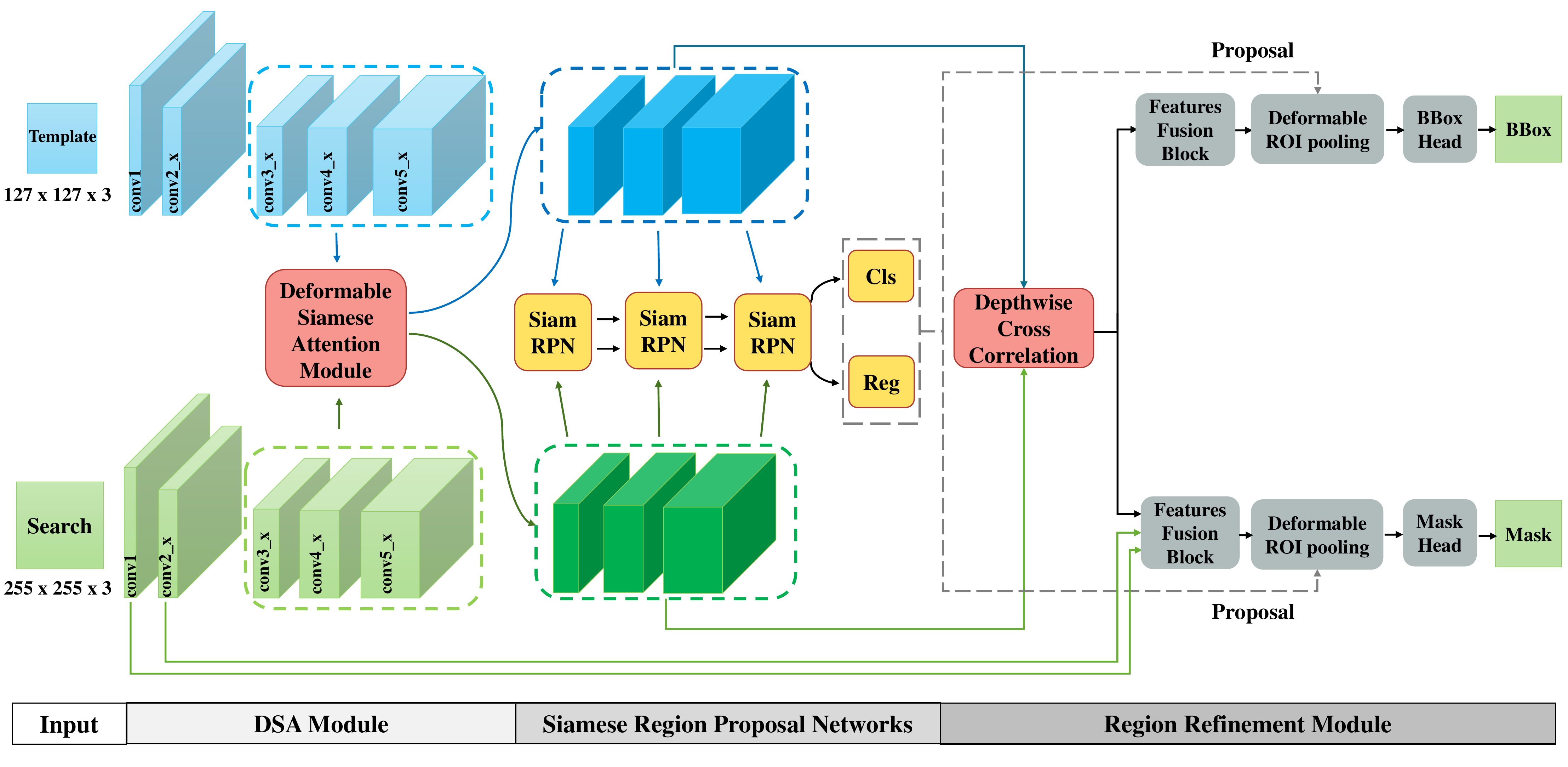}
		\end{center}
		\vspace*{-3mm} 
		\caption{An overview of the proposed Deformable Siamese Attention Networks (SiamAttn). It consists of a deformable Siamese attention (DSA) module, Siamese region proposal networks (SiamRPN) and a region refinement module. The features of the last three stages are extracted and then modulated by the DSA module. It generates two-stream attentional features which are fed into SiamRPN blocks to predict a single tracking region, further refined by the refinement module.  }
		\vspace{-10pt}
		\label{fig:network-arc}
	\end{figure*}
	
	However, Siamese-based trackers can be affected by distractors with complex backgrounds.
	Recent work attempted to design various strategies to update the template online, in an effort to improve the  target-discriminability of Siamese-based trackers, such as MLT~\cite{choi2019deep}, UpdateNet~\cite{zhang2019learning} and GradNet~\cite{li2019gradnet}. An alternative solution is to extend existing online discriminative framework with deep networks for end-to-end learning, e.g., ATOM~\cite{danelljan2019atom} and DiMP~\cite{bhat2019learning}. 
	In addition, Zhu \etal~\cite{zhu2018end} exploited motion information in Siamese networks to improve the feature representation. 
	
	Recently, the attention mechanism has been widely applied in various tasks. 
	Hu \etal~\cite{hu2018squeeze} proposed a SENet to enhance the representational power of the networks by modeling channel-wise relationships via attentions. 
	Wang \etal~\cite{wang2018non} developed a non-local operation in the space-time dimension to guide the aggregation of contextual information. In~\cite{fu2019dual}, a self-attention mechanism  was introduced to harvest the contextual information for semantic segmentation. 
	Particularly, Wang \etal~\cite{wang2018learning} proposed a RASNet by developing an attention mechanism for Siamese trackers, but it only utilizes the template information, which might limit its representation ability.
	To better explore the potentials of feature attentions in Siamese networks, we compute both self-attention and cross-branch attention jointly with deformable operations to enhance the discriminative representation of target.

	\section{Deformable Siamese Attention Networks}\label{sec:method}
	
	We describe the details of our Deformable Siamese Attention Networks (SiamAttn). As shown in Fig.~\ref{fig:network-arc}, it consists of three main components: a deformable Siamese attention (DSA) module, Siamese region proposal networks (Siamese RPN), and a region refinement module. \\

	\noindent\textbf{Overview.}
	We use a five-stage ResNet-50 as the backbone of Siamese networks, which computes increasingly high-level features as the layers become deeper.
	The features of the last three stages on both Siamese branches can be modulated and enhanced by the proposed DSA module, generating two-stream attentional features.
	%
	Then we apply three Siamese RPN blocks described in \cite{li2019siamrpn++} to the attentional features, generating dense response maps, which are further processed by a classification head and a bounding box regression head to predict a single tracking region. Finally, the generated tracking region is further refined by a region refinement module, where
	depth-wise cross correlations are computed on the two-stream attentional features. The correlated features are further fused and enhanced, and then are used for refining the tracking region via joint bounding box regression and target mask prediction.
	
	\begin{figure*}
		\small
		\begin{center}
			\includegraphics[width=0.95\linewidth]{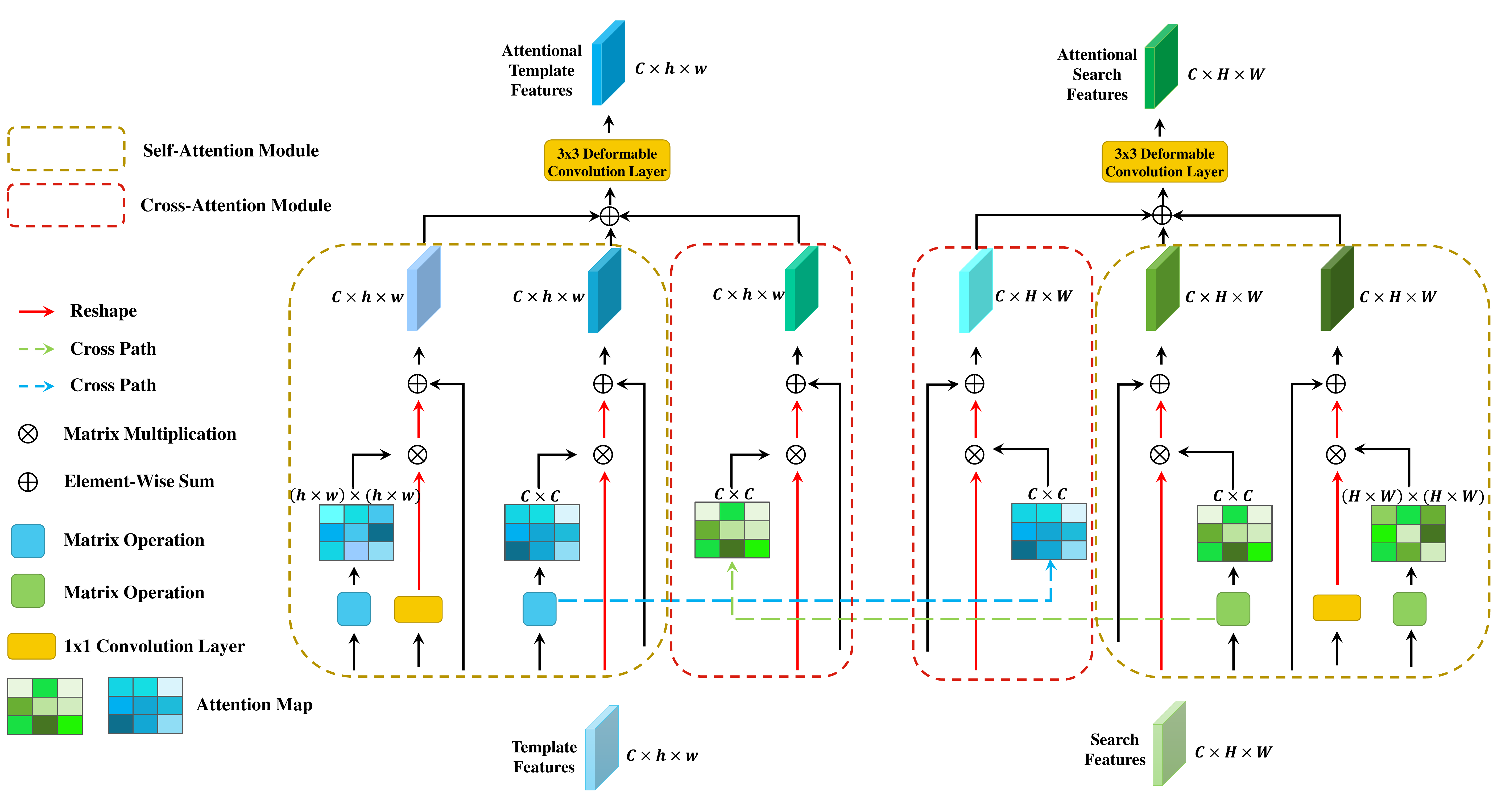}
		\end{center}
		\vspace{-5mm}
		\caption{The proposed Deformable Siamese Attention (DSA) module, which consists of two sub-modules: self-attention sub-module and cross-attention sub-module.  It takes template features and search features as inputs, and computes corresponding attentional features. The self-attention can learn strong context information via spatial attention, and at the same time, selectively emphasizes interdependent channel-wise features with channel attention. The cross-attention aggregates rich contextual interdependencies between the target template and the search image. }
		\vspace{-10pt}
		\label{fig:attention}
	\end{figure*}
	
	\subsection{Siamese-based Trackers}
	Bertinetto \etal \cite{bertinetto2016fully} introduced Siamese networks for visual object tracking, 
	which formulates visual object tracking as a similarity learning problem.
	Siamese networks consist of a pair of CNN branches with sharing parameters $\phi$, which are used to project the target image ($z$) and the search image ($x$) into a common embedding space, where a similarity metric $g$ can be computed to measure the similarity between them, $g(\phi(x), \phi(z))$.
	%
	Li \etal~\cite{li2018high} applied region proposal networks (RPN)~\cite{girshick2015fast} with Siamese networks for visual object tracking (referred as SiamRPN), where the computed features $\phi(x)$ and $\phi(z)$ are fed into the RPN framework using an up-channel cross-correlation operation. This generates dense response maps where RPN-based detection can be implemented, leading to significant performance improvements.
	
	%

	\noindent\textbf{SiamRPN++.} In~\cite{li2019siamrpn++}, SiamRPN++ was introduced to improve the performance of SiamRPN, by exploring the power of deeper networks.
	A spatial aware sampling strategy was developed to address a key limitation of the Siamese-based trackers, allowing them to benefit from a deeper backbone likes ResNet-50.
	Furthermore, SiamRPN++ adopts a depth-wise cross correlation to replace the up-channel cross correlation, which reduces the number of parameters and accelerates the training process.
	Moreover, it aggregates  multi-layer features to predict the target more accurately.
	%
	Similarly, we use ResNet-50 as backbone, with depth-wise cross correlation and multi-layer aggregation strategy, by following SiamRPN++~\cite{li2019siamrpn++}. But we introduce a new Siamese attention module that enhances the learned discriminative representations of the target object and the search image, which is the key to improve the tracking performance on both accuracy and robustness.
	
	\subsection{Deformable Siamese Attention Module}\label{sdam}
	
	As illustrated in 
	Fig.~\ref{fig:attention}, the proposed DSA module takes a pair of convolutional features computed from Siamese networks as inputs, and outputs the modulated features by applying the Siamese attention mechanism.
	The DSA module consists of two sub-modules:
	a \textit{self-attention} sub-module and a \textit{cross-attention} sub-module.
	We denote the feature maps of the target and the search image as $\textbf{Z}$ and $\textbf{X}$, with feature shapes of $C \times h \times w$ and $C \times H \times W$.
	
	\noindent\textbf{Self-Attention.}
	Inspired by~\cite{fu2019dual}, our self-attention sub-module attends to two aspects, namely channels and special positions.
	Unlike the classification or detection task where object classes are pre-defined,
	visual object tracking is a class-agnostic task and the class of object is fixed during the whole tracking process.
	As observed in~\cite{li2019siamrpn++}, each channel map of the high-level convolutional features usually responses for a specific object class.
	Equally treating the features across all channels will hinder the representation ability.
	Similarly, as limited by receptive fields, the features computed at each spatial position of the maps can only capture the information from a local patch.
	Therefore, it is crucial to learn the global context from the whole image.
	
	Specifically, self-attention is computed separately on the target branch and the search branch, and both channel self-attention and spatial self-attention are calculated at each branch. 
	Taking the spatial self-attention for example. Suppose the input features are $\textbf{X} \in \mathbb{R}^{C \times H \times W}$, we first apply two separate convolution layers with $1 \times 1$ kernels on $\textbf{X}$ to generate query features $\textbf{Q}$ and key features $\textbf{K}$ respectively,
	where $\textbf{Q},\textbf{K} \in \mathbb{R}^{C^{\prime} \times H \times W}$ and $C^{\prime} = \frac{1}{8} C$ is the reduced channel number.
	The two features are then reshaped to $\bar{\textbf{Q}},\bar{\textbf{K}} \in \mathbb{R} ^ {C^{\prime} \times N}$ where $N = H \times W$.
	We can generate a spatial self-attention map $\textbf{A}^{\mbox{s}}_{\mbox{s}} \in \mathbb{R}^{N \times N}$
	via matrix multiplication and column-wise softmax operations as,
	\begin{equation}
	\textbf{A}^{\mbox{s}}_{\mbox{s}} = \mbox{softmax}_{col}(\bar{\textbf{Q}}^T  \bar{\textbf{K}}) \in \mathbb{R}^{N \times N}.
	\end{equation}
	Meanwhile, a 1$\times$1 convolution layer with a reshape operation is applied to the features $\textbf{X}$ to generate value features $\bar{\textbf{V}} \in \mathbb{R}^{C \times N}$, which are multiplied with the attention map and then are added to the reshaped features $\bar{\textbf{X}} \in \mathbb{R}^{C \times N}$ with a residual connection as,
	\begin{equation}
	\bar{\textbf{X}}^{\mbox{s}}_{\mbox{s}} = \alpha \bar{\textbf{V}} \textbf{A}^{\mbox{s}}_{\mbox{s}} + \bar{\textbf{X}} \in \mathbb{R}^{C \times N}.
	\end{equation}		
	where $\alpha$ is a scalar parameter.
	The outputs are then reshaped back to the original size as $\textbf{X}^{\mbox{s}}_{\mbox{s}} \in \mathbb{R}^{C \times H \times W}$.
	
	We can compute channel self-attention $\textbf{A}^{\mbox{s}}_{\mbox{c}}$ 
	and the channel-wise attentional features $\textbf{X}^{\mbox{s}}_{\mbox{c}}$ in a similar manner. Notice that on computing the channel self-attention and the corresponding attentional features, the query, key and value features are the original convolutional features computed directly from the Siamese networks, without implementing $1\times 1$ convolutions. The final self-attentional features  $\textbf{X}^{\mbox{s}}$ are generated by simply combining the spatial and channel-wise attentional features using element-wise sum. 
	\\
	
	\begin{figure}[t]
		\small
		\begin{center}
			\includegraphics[width=1.0\linewidth]{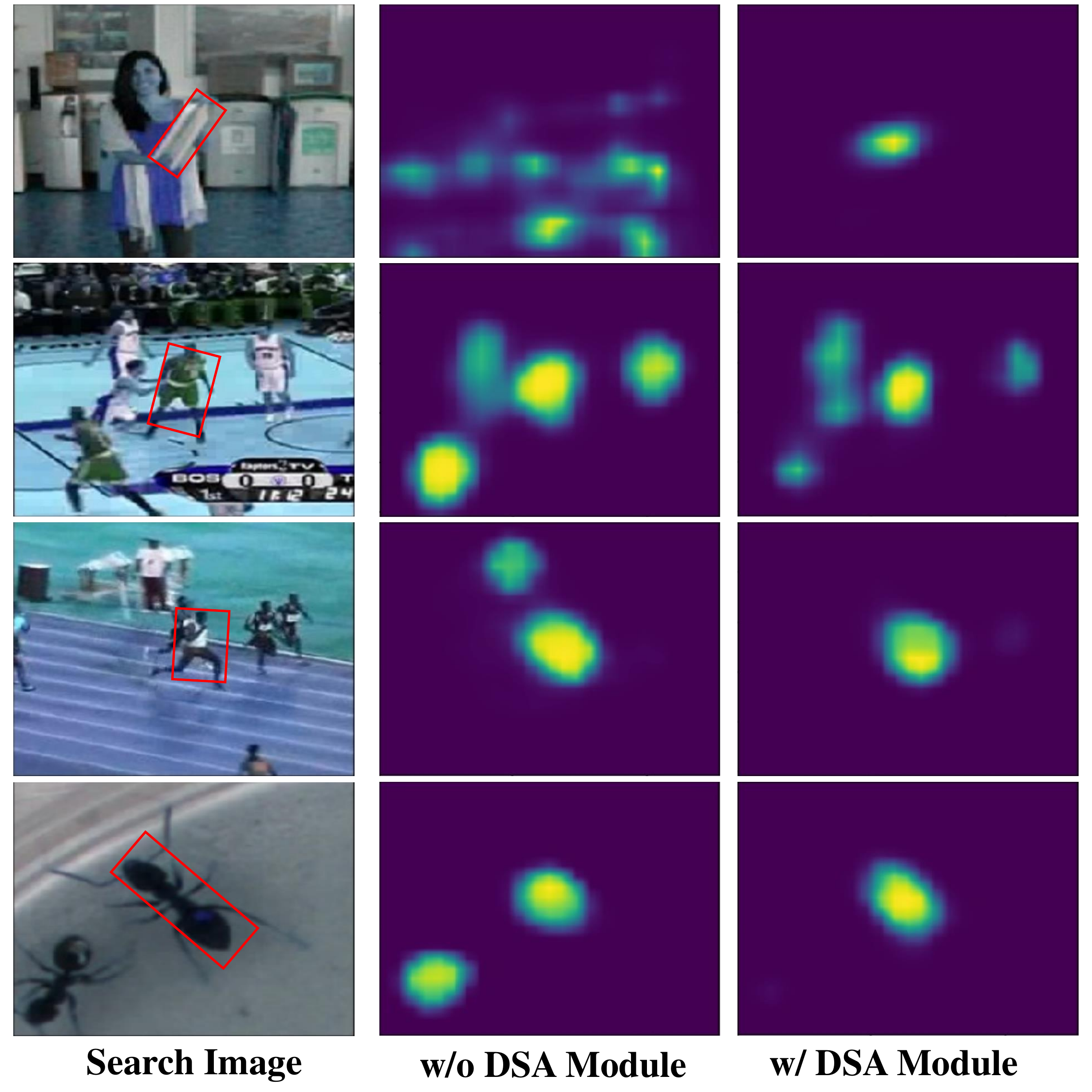}
		\end{center}
		\vspace{-4mm}
		\caption{Visualization of confidence maps. The $1^{st}$ column: search images, the $2^{nd}$ column: confidence maps without our DSA  module, and the $3^{rd}$ column: confidence maps with DSA module which enhances target-background discriminability in the computed attentional features.}
		\vspace{-10pt}
		\label{fig:attention-examples}
	\end{figure}	
	
	\noindent\textbf{Cross-Attention.}
	Siamese networks usually make predictions in the last stage,
	while the features from two branches are computed separately, but may compensate each other. It is common that multiple objects appear at the same time, even with occlusions during tracking.
	%
	%
	Therefore, it is of great importance for the search branch to learn  the target information, which enables it to generate a more discriminative representation that helps identify the target more accurately.
	Meanwhile, the target representation can be more meaningful when the contextual information from the search image is encoded. 
	To this end, we propose a cross-attention sub-module to learn such mutual information from two Siamese branches, which in turn allows the two branches to work more collaboratively. 
	
	Specifically, we use
	$\textbf{Z} \in \mathbb{R}^{C\times h \times w}$ and $\textbf{X} \in \mathbb{R}^{C\times H \times W}$ to denote template features and search features, respectively.
	Taking the search branch for example, we first reshape the target features $\textbf{Z}$ to $\bar{\textbf{Z}} \in \mathbb{R}^{C \times n}$  where $n = h \times w$.
	Then we compute the cross-attention from the target branch by performing similar operations as channel self-attention,
	\begin{equation}
	\textbf{A}^{\mbox{c}} = \mbox{softmax}_{row}(\bar{\textbf{Z}} \bar{\textbf{Z}}^T ) \in \mathbb{R}^{C \times C}.	
	\end{equation}
	where row-wise softmax is implemented on the computed matrix.
	Then the cross-attention computed from the target branch is encoded into the search features $\textbf{X}$ as,
	\begin{equation}
	\bar{\textbf{X}}^{\mbox{c}} = \gamma \textbf{A}^{\mbox{c}} \bar{\textbf{X}} + \bar{\textbf{X}} \in \mathbb{R}^{C \times N}.
	\end{equation}
	where $\gamma$ is a scalar parameter, and the reshaped features $\textbf{X}^{\mbox{c}} \in \mathbb{R}^{C \times H \times W}$ are the output of the sub-module.
	
	Finally, the self-attentional features $\textbf{X}^{\mbox{s}}$  and the cross-attentional features $\textbf{X}^{\mbox{c}}$ are simply combined with an element-wise sum, generating the attentional features for the search image. The attentional features for target image can be computed in a similar manner. \\
	
	\noindent\textbf{Deformable Attention.} 
	The building units in CNNs, such as convolution or pooling units, often have fixed geometric structures, by assuming the objects are rigid. 
	For object tracking, it is of importance to model complex geometric transformations
	because the tracking objects usually have large deformations due to various factors, such as viewpoint, pose, occlusion and so on. 
	The proposed attention mechanism can handle such challenges to some extent.  We further introduce deformable attention to enhance the capability for handling such  geometric transformations.
	%
	
	The deformable attention can sample the input feature maps at variable locations instead of the fixed ones, making them attend to the content of objects with deformations. 
	Therefore, it is particularly suitable for object tracking, where the visual appearance of a target can be changed significantly over time.  
	Specifically, a $3 \times 3$ deformable convolution~\cite{dai2017deformable} is further applied to the computed attentional features, generating the final attentional features which are more accurate, discriminative and robust. 
	As shown in Fig.~\ref{fig:attention-examples}, with our DSA module, the confidence maps of the attentional features focus more accurately on the interested objects, making the objects more discriminative against distractors and background. \\
	
	\noindent\textbf{Region Proposals.}
	The DSA module outputs Siamese attentional features for both target image and search image. Then we apply three Siamese RPN blocks on the attentional features for generating a set of target proposals, with corresponding bounding boxes and class scores, as shown in Fig.~\ref{fig:network-arc}. Specifically, a Siamese RPN block is a combination of multiple fully convolutional layers, depth-wise cross correlation, with a regression head and a classification head on top, as described in~\cite{li2018high}. It takes a pair of convolutional features computed from the two branches of Siamese networks, and outputs dense prediction maps. By following~\cite{li2019siamrpn++}, we apply three Siamese RPN blocks for the Siamese features computed from the last three stages, generating three prediction maps which are further combined by a weighted sum.
	Each spatial position of the combined maps predicts a set of region proposals, corresponding to the pre-defined anchors. Then the predicted proposal with the highest classification score is selected as the output tracking region.
	
	\subsection{Region Refinement Module}
	We further develop a region refinement module to improve the localization accuracy of the predicted target region.
	We first apply depth-wise cross correlations between the two attentional features across multiple stages, generating multiple correlation maps. Then the correlation maps are fed into a fusion block, where the feature maps with different sizes are aligned in both spatial and channel domains, e.g., by using up-sampling or down-sampling, with  $1 \times 1$ convolution. Then the aligned features are further fused (with element-wise sum) for predicting both bounding box and mask of the target. Besides, we further perform two additional operations: (1) we combine the convolutional features of the first two-stages into the fused features, which encodes richer local detailed information for mask prediction; (2) a deformable RoI pooling~\cite{dai2017deformable} is applied to compute target features more accurately. Bounding box regression and mask prediction often require different levels of convolutional features. Thus we generate the convolutional features with spatial resolutions of  $64 \times 64$ for mask prediction and $25 \times 25$ for bounding box regression.

	Notice that the classification head is not applied since visual object tracking is a class-agnostic task.
	The input resolution for the bounding box head is $4 \times 4$.
	By using  two fully-connected layers with 512 dimensions,
	the bounding box head predicts a 4-tuple $t = (t_{x}, t_{y}, t_{w}, t_{h})$.
	Similarly, the input of mask prediction head has a  spatial resolution of $16 \times 16$.
	By using four convolutional layers and a de-convolutional layer, the mask head predicts a class-agnostic binary mask with 64 $\times$ 64 for the tracking object.
	Compared to ATOM~\cite{danelljan2019atom} and SiamMask~\cite{wang2019fast} which predict bounding boxes and masks
	\emph{densely}, our refinement module uses light-weight convolutional heads to predict a bounding box and a mask for a single tracking region, which is much more computationally efficient.
	\subsection{Training Loss}
	
	Our model is trained in an end-to-end fashion, where the training loss is a weighted combination of multiple functions from Siamese RPN and region refinement module:
	\begin{equation}
	\begin{aligned}
	L = & L_{\mbox{rpn-cls}} + \lambda_{1}L_{\mbox{rpn-reg}} + \\
	& \lambda_{2} L_{\mbox{refine-box}} + \lambda_{3} L_{\mbox{refine-mask}}.
	\end{aligned}
	\end{equation}
	where $L_{\mbox{rpn-cls}}$ and $L_{\mbox{rpn-reg}}$ refer to a classification loss and a regression loss in Siamese RPN.
	We employ a negative log-likelihood loss and a smooth L1 loss correspondingly.
	Similarly, $L_{\mbox{refine-box}}$ and $L_{\mbox{refine-mask}}$ indicate a smooth L1 loss for bounding box regression and a binary cross-entropy loss for mask segmentation in region refinement module.
	The weight parameters $\lambda_{1}$, $\lambda_{2}$ and $\lambda_{3}$ are used to balance different tasks, 
	which are empirically set to 0.2, 0.2 and 0.1 in our implementation.
	
	
	\section{Experiments and Results}\label{sec:exp}
	We conduct extensive experiments on six benchmark databases: OTB-2015 \cite{wu2013online}, 
	UAV123 \cite{mueller2016benchmark}, VOT2016 \cite{kristan2016visual}, VOT2018 \cite{kristan2018sixth}, LaSOT \cite{fan2019lasot} and TrackingNet \cite{muller2018trackingnet} datasets 
	and provide ablation study to verify the effects of each proposed component. 
	
	\subsection{Datasets}
	\noindent\textbf{OTB-2015~\cite{wu2013online}.} OTB-2015 is one of the most commonly used benchmarks for visual object tracking. It has 100 fully annotated video sequences using two evaluation metrics, a precision score and an area under curve (AUC) of success plot.
	The precision score is the percentage of frames in which the distance between the center of tracking results and ground-truth is under 20 pixels. 
	The success plot shows the ratios of successfully tracked frames with various thresholds.
	
	\noindent\textbf{VOT2016~\cite{kristan2016visual} \& VOT2018~\cite{kristan2018sixth}.} VOT2016 and VOT2018 are widely-used benchmarks for visual object tracking.
	VOT2016 contains 60 sequences with various challenging factors while VOT2018 has 10 different sequences with VOT2016. 
	The two datasets are annotated with the rotated bounding boxes, and a reset-based methodology is applied for evaluation.
	For both benchmarks, trackers are measured in terms of
	accuracy (A), robustness (R), and expected average overlap (EAO).
	
	\noindent\textbf{UAV123~\cite{mueller2016benchmark}.} UAV123 
	contains 123 sequences captured from low-altitude UAVs.
	Unlike other tracking datasets, the viewpoint of UAV123 is aerial and the targets to be tracked are usually small.
	
	\noindent\textbf{LaSOT~\cite{fan2019lasot}.} LaSOT 
	is a large-scale dataset with 1400 sequences in total, and 280 sequences in test set.
	high-quality dense annotations are provided, and deformation and occlusion are very common in LaSOT. The average sequence length of LaSOT is 2500 frames, demonstrating long-term performance of the evaluated trackers.
	
	\noindent\textbf{TrackingNet~\cite{muller2018trackingnet}.} TrackingNet
	contains 30000 sequences with 14 million dense annotations and 511 sequences in the test set.
	It covers diverse object classes and scenes, 
	requiring trackers to have both discriminative and generative capabilities.
	
	\subsection{Implementation Details}\label{sec:impdetail}
	
	We use  ResNet-50, pre-trained on ImageNet~\cite{deng2009imagenet}, as the backbone, and the whole networks are then fine-tuned on the training sets of COCO~\cite{lin2014microsoft}, 
	YouTube-VOS~\cite{xu2018youtube}, LaSOT~\cite{fan2019lasot} and TrackingNet~\cite{muller2018trackingnet}.
	We apply stochastic gradient descent (SGD) with a momentum of 0.9 and a weight decay of $10^{-5}$ for optimization.
	By following SiamFC~\cite{bertinetto2016fully}, we adopt an exemplar image of $127 \times 127$ and a search image of $255\times 255$ for training and testing. 
	
	Our model is trained for 20 epochs. 
	By following SiamRPN++~\cite{li2019siamrpn++},
	we use a warm-up learning rate of $10^{-3}$ for the first 5 epochs 
	which decays exponentially from $5\times10^{-3}$ to $5\times10^{-4}$ for the last 15 epochs.
	The weights of backbone are frozen
	for the first 10 epochs, then the whole networks are trained end-to-end for the last 10 epochs. 
	In particular, the learning rate of backbone is 20 times smaller than the other parts.
	Batch size is set to 12. 
	Our anchor boxes have 5 aspect ratios, $[0.33, 0.5, 1, 2, 3]$.  
	In Siamese-RPN blocks, an anchor box is labelled as positive when it has an IoU$>0.6$, or as negative when IoU$<0.3$. 
	Other patches whose IoU overlap falls in between are ignored. 
	In addition, we sample 16 regions from each image with IoU$>0.5$ for training our region refinement module.
	
	For the backbone networks, we employ dilated convolutions for the last two blocks to increase the receptive fields, and the effective strides at these two blocks are reduced from 16 or 32 pixels to 8 pixels. We also reduce the number of feature channels to 256 for the last three blocks of the backbone networks via a 1 $\times$ 1 convolution layer.
	During inference, cosine window penalty, scale change penalty and linear interpolation update strategy~\cite{li2018high} are applied. 
	Only one single region with the highest score  predicted  by Siamese RPN blocks is fed into our region refinement module. 
	Our method is implemented using PyTorch, and we use NVIDIA GeForce RTX 2080Ti GPU.
	
	
	\begin{figure}[tbp]
		\begin{subfigure}[b]{0.23\textwidth}
			\includegraphics[width=1\textwidth]{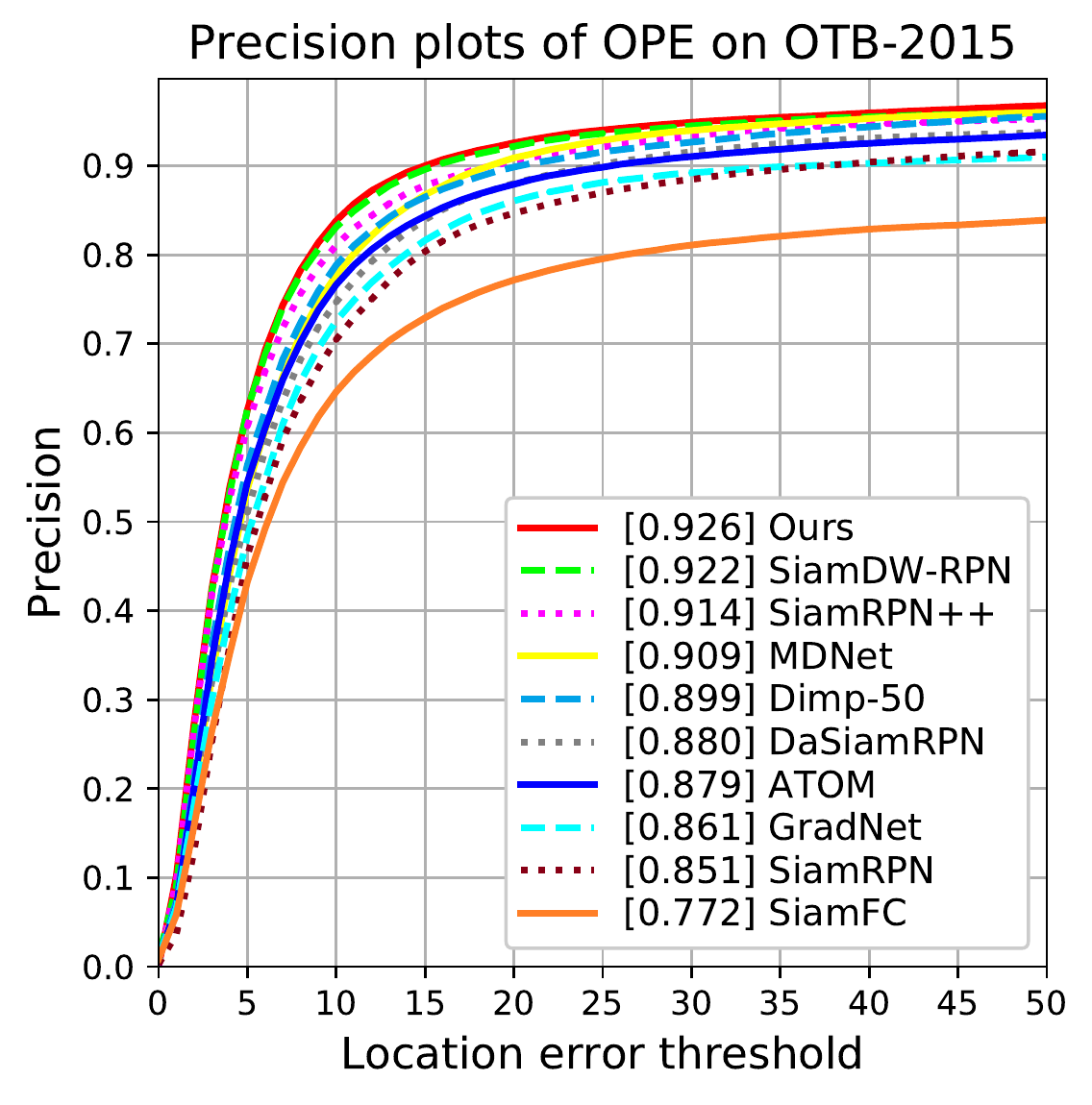}
			\caption{Precision Plot}
			\label{fig:otb-a}
		\end{subfigure}
		\begin{subfigure}[b]{0.23\textwidth}
			\includegraphics[width=1\textwidth]{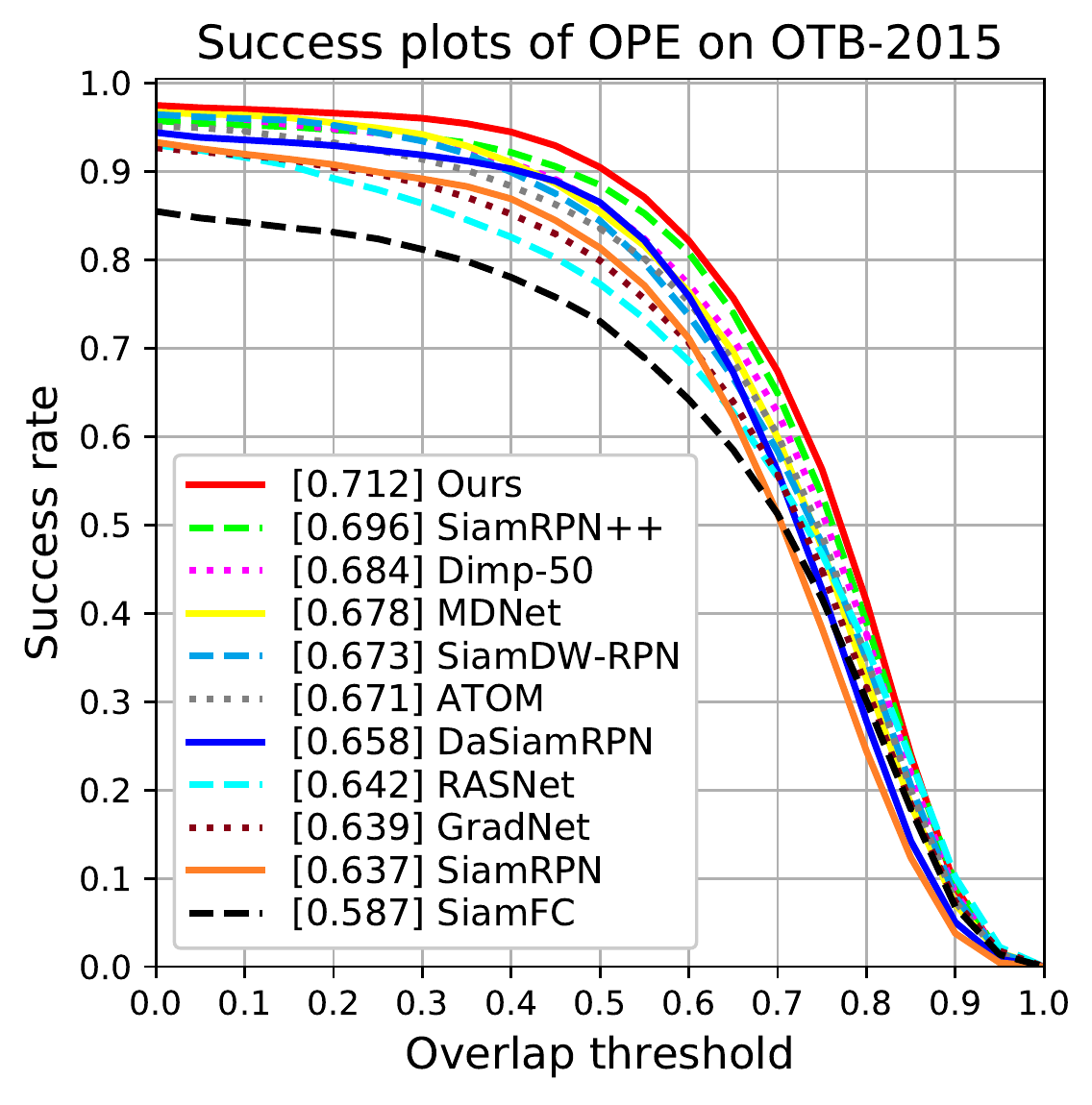}
			\caption{Success Plot}
			\label{fig:otb-b}
		\end{subfigure}
		\caption{Comparisons with state-of-the-art methods on success and precision plots on OTB-2015 dataset.}
		\vspace{-10pt}
		\label{Fig:OTB}
	\end{figure}
	
	\subsection{Comparisons with the State-of-the-art}
	
	
	\noindent\textbf{On OTB-2015.}
	Fig.~\ref{Fig:OTB} shows quantitative results on OTB-2015
	dataset. 
	Our tracker achieves the best AUC and precision score
	among all methods for this widely studied dataset.
	Specifically, we obtain a precision of 0.712 and
	an AUC of 0.926 which surpass that of SiamRPN++~\cite{li2019siamrpn++} by
	1.6\%  and 1.2\%, respectively. 
	
	\noindent\textbf{On VOT2016 \& VOT2018.} 
	The results on VOT2016 and VOT2018 are compared in Tab.~\ref{Tab:VOT}. 
	Our tracker achieves 0.68 accuracy, 0.14 robustness and 0.537 EAO on VOT2016, outperforming the state-of-the-art methods under all metrics. 
	Compared with recent SiamRPN++~\cite{li2019siamrpn++} and SiamMask~\cite{wang2019fast}, our method has significant improvements of 7.3\% and 9.5\% on EAO respectively.
	On VOT2018, our method achieves the top EAO score while having competitive accuracy and robustness with other state-of-the-art methods. 
	SiamMask-\textit{Opt}~\cite{wang2019fast} attains the best accuracy by finding the optimal rotated rectangle from a binary mask, which however increases the computational cost significantly, and reduces its fps to $5$. 
	Our method only uses a single rotated minimum bounding rectangle from the predicted mask, which achieves a comparable accuracy of $0.63$, but has a large improvement on EAO, 0.387$\rightarrow$0.470, with a real-time running speed at $33$ fps. 
	Compared with SiamRPN++ and recent leading tracker Dimp-50~\cite{bhat2019learning}, our tracker obtains a clear performance gain of 5.5\% and 3.0\% respectively in terms of EAO,
	demonstrating the efficiency of the proposed Siamese attention and refinement module. 
	
	\begin{table}
		\setlength{\tabcolsep}{3pt}
		\begin{center}
			\resizebox{\columnwidth}{!}{%
				\begin{tabular}{*{12}{l c c c c c c}}
					\toprule 
					\multirow{2.5}*{Tracker} & \multicolumn{3}{c}{VOT2016} & \multicolumn{3}{c}{VOT2018}  \\
					\cmidrule(lr){2-4} \cmidrule(lr){5-7} 
					& A$\uparrow$ & R$\downarrow$ & EAO$\uparrow$ & A$\uparrow$ & R$\downarrow$ & EAO$\uparrow$	\\
					\midrule
					SiamFC~\cite{bertinetto2016fully}    & 0.53 & 0.46 & 0.235 & 0.50 & 0.59 & 0.188  \\
					MDNet~\cite{nam2016learning}         & 0.54 & 0.34 & 0.257 & - & - & -  \\
					C-COT~\cite{danelljan2016beyond}     & 0.54 & 0.24 & 0.331 & 0.49 & 0.32 & 0.267  \\
					FlowTrack~\cite{zhu2018end}          & 0.58 & 0.24 & 0.334 & - & - & - \\
					SiamRPN~\cite{li2018high}            & 0.56 & 0.26 & 0.344 & - & - & -  \\
					C-RPN~\cite{fan2019siamese}          & 0.59 & - & 0.363 & - & - & -  \\
					ECO~\cite{danelljan2017eco}          & 0.55 &{\color{blue}0.20}& 0.375 & 0.48 & 0.28 & 0.276  \\
					DaSiamRPN~\cite{zhu2018distractor}   & 0.61 & 0.22 & 0.411 & 0.59 & 0.28 & 0.383  \\
					SPM~\cite{wang2019spm}               & 0.62 & 0.21 & 0.434 & - & - & - \\
					SiamMask-\textit{Opt}~\cite{wang2019fast}&{\color{blue}0.67} & 0.23 & 0.442 &{\color{red}0.64} & 0.30 & 0.387  \\
					UpdateNet~\cite{zhang2019learning}   & 0.61 & 0.21 &{\color{blue}0.481} & - & - & 0.393  \\
					GFS-DCF~\cite{xu2019joint}           & - & - & - & 0.51 &{\color{red}0.14}& 0.397  \\
					ATOM~\cite{danelljan2019atom}        & - & - & - & 0.59 & 0.20 & 0.401  \\
					SiamRPN++~\cite{li2019siamrpn++}     & 0.64 &{\color{blue}0.20}&0.464& 0.60 & 0.23 & 0.415  \\
					Dimp-50~\cite{bhat2019learning}      & - & - & - & 0.60 &{\color{blue}0.15} & {\color{blue}0.440}  \\
					\midrule
					Ours   &{\color{red}0.68}&{\color{red}0.14}&{\color{red}0.537}&{\color{blue}0.63}&0.16&{\color{red}0.470} \\
					\bottomrule
				\end{tabular}
			}
		\end{center}
		\vspace*{-4mm} 
		\caption{Results on  VOT2016 and VOT2018, with accuracy (A), robustness (R) and expected average overlap (EAO).}
		\vspace{-10pt}
		\label{Tab:VOT}
	\end{table}
	
	\noindent\textbf{On UAV123.} 
	As shown in Tab.~\ref{Tab:UAV}, SiamAttn attains the best precision, improving the closest  one: SiamRPN++, from  $0.807$ to  $0.845$, while having a comparable AUC with  DiMP-50 which did not report the precision score.
	
	\begin{table} 
		\begin{center}
			\resizebox{\columnwidth}{!}{%
				\setlength{\tabcolsep}{1pt}{
					\begin{tabular}{*{10}{c}}
						\toprule 
						& ARCF & ECO & SiamRPN & DaSiam- & SiamRPN++ & ATOM & Dimp-50 & Ours	\\
						&~\cite{huang2019learning}&~\cite{danelljan2017eco}&~\cite{li2018high}&RPN~\cite{zhu2018distractor} 
						&~\cite{li2019siamrpn++}&~\cite{danelljan2019atom}&~\cite{bhat2019learning}&      \\
						\midrule
						AUC & 0.47 & 0.525 & 0.527 & 0.586 & 0.613 & 0.644 & {\color{red}0.654} &{\color{blue}0.650}    \\
						Pr  & 0.67 & 0.741 & 0.748 & 0.796 &{\color{blue}0.807} & - & - &{\color{red}0.845}    \\
						
						\bottomrule
					\end{tabular}
				}
			}
		\end{center}
		\vspace*{-4mm} 
		\caption{Results on UAV123.}
		\vspace{-10pt}
		\label{Tab:UAV}
	\end{table}
	
	\noindent\textbf{On LaSOT.} 
	Tab.~\ref{Tab:LaSOT} shows the comparison results on LaSOT with long sequences. 
	Our method attains the best normalized precision, outperforming SiamRPN++ considerably by 56.9\%$\rightarrow$64.8\% (with 49.5\%$\rightarrow$56.0\% on success). 
	Again, our method has a comparable success score with DiMP-50, while attaining a higher normalized precision. 
	
	\begin{table} 
		\begin{center}
			\resizebox{\columnwidth}{!}{%
				\setlength{\tabcolsep}{.4mm}{
					\begin{tabular}{*{9}{c}}
						\toprule 
						& MLT &MDNet & DaSiam- & Update- & SiamRPN++ & ATOM & Dimp-50 & Ours	\\
						&~\cite{choi2019deep}&~\cite{nam2016learning}&RPN~\cite{zhu2018distractor} 
						&Net~\cite{zhang2019learning}&~\cite{li2019siamrpn++}&~\cite{danelljan2019atom}
						&~\cite{bhat2019learning}&\\
						\midrule
						Success(\%) & 34.5 & 39.7 & 41.5 & 47.5 & 49.5 & 51.5 &{\color{red}56.9}&{\color{blue}56.0} \\
						Norm.Pr(\%) & - & 46.0 & 49.6 & 56.0& 56.9& 57.6& {\color{blue}64.3}&{\color{red}64.8}  \\
						
						\bottomrule
					\end{tabular}
				}
			}
		\end{center}
		\vspace*{-4mm} 
		\caption{Results on LaSOT.}
		\vspace{-10pt}
		\label{Tab:LaSOT}
	\end{table}
	
	\noindent\textbf{On TrackingNet.} 
	We further evaluate SiamAttn on the large-scale TrackingNet. 
	As illustrated in Tab.~\ref{Tab:Tracking}, it outperforms all previous methods consistently.
	Compared with the most recent DiMP-50, SiamAttn has improvements of 1.2\% on success, and  1.6\% on normalized precision,
	demonstrating its ability to handle diverse objects over complex scenes. 
	
	\begin{table} 
		\begin{center}
			\resizebox{\columnwidth}{!}{%
				\setlength{\tabcolsep}{1pt}{
					\begin{tabular}{*{9}{c}}
						\toprule 
						& GFS- & DaSiam- & Update- & ATOM &SPM & SiamRPN++ & Dimp-50 & Ours	\\
						&DCF~\cite{xu2019joint}&RPN~\cite{zhu2018distractor}
						& Net~\cite{zhang2019learning} &~\cite{danelljan2019atom} &~\cite{wang2019spm} 
						&~\cite{li2019siamrpn++}&~\cite{bhat2019learning}&\\
						\midrule
						Success(\%)    & 60.9 & 63.8 & 67.7 & 70.3 &71.2 & 73.3  &{\color{blue}74.0}&{\color{red}75.2}\\
						Norm.Pr(\%)    & 71.2 & 73.3 & 75.2 & 77.1 &77.8 & 80.0  &{\color{blue}80.1}&{\color{red}81.7}\\
						
						\bottomrule
					\end{tabular}
				}
			}
		\end{center}
		\vspace*{-4mm} 
		\caption{Results on TrackingNet.}
		\vspace{-10pt}
		\label{Tab:Tracking}
	\end{table}
	
	\subsection{Ablation Study}
	We study  the impact of individual components in SiamAttn, and conduct 
	ablation study on VOT2016. \\
	
	\noindent\textbf{Model Architecture. } 
	We use SiamRPN++~\cite{li2019siamrpn++} as baseline.  As shown in Tab.~\ref{Tab:ablation}, SiamRPN++ achieves an EAO of $0.464$. 
	By adding mask learning layers to SiamRPN++, the EAO can be improved to 0.477.
	With our region refinement module, the EAO score is further increased by $+2.2\%$.
	Compared with the baseline, the accuracy score improves from $0.64$ to $0.67$, with comparable robustness. 
	Our Siamese attention consists of both self-attention and cross-attention, each of which can further improve the EAO  by $+4.7\%$ or $+4.9\%$ respectively.
	This suggests that the proposed cross-attention is critical to the tracking results, and even has a more significant impact than the self-attention.
	Jointly exploring both self-attention and cross-attention makes our method not only robust, but also more accurate. This results in a high EAO of $0.537$, surpassing the baseline  by a large margin of $7.3\%$. \\
	
	\begin{table}
		\setlength{\tabcolsep}{3.2pt}
		\begin{center}
			\resizebox{\columnwidth}{!}{%
				\begin{tabular}{*{14}{l c c c c}}
					\toprule 
					Method &  A$\uparrow$ & R$\downarrow$ & EAO$\uparrow$ & $\Delta$EAO \\
					\midrule
					Baseline                   & 0.64 & 0.20 & 0.464 & - \\
					Baseline+ML                & 0.66 & 0.21 & 0.477 & +1.3\%\\
					Baseline+RR                & 0.67 & 0.19 & 0.486 & +2.2\%\\
					Baseline+RR+SA              & 0.66 & 0.16 & 0.511 & +4.7\%\\              
					Baseline+RR+CA              & 0.67 & 0.15 & 0.513 & +4.9\%\\             
					\midrule
					Baseline+RR+CA+SA (Ours)   & 0.68 & 0.14 & 0.537 & +7.3\%\\             
					
					\bottomrule
				\end{tabular}
			}
		\end{center}
		\vspace*{-4mm} 
		\caption{Ablation study on VOT2016. SiamRPN++ is baseline. ML: Mask Learning, RR: Region Refinement (including ML), SA: Self-Attention, and CA: Cross-Attention. }
		\vspace{-10pt}
		\label{Tab:ablation}
	\end{table}
	
	\noindent\textbf{Deformable Layers. } 
	In this study, we evaluate the impact of deformable operations by replacing them with a regular convolution.
	As shown in Tab.~\ref{Tab:deform}, this results in slight performance drops, e.g., 	0.537$\rightarrow$0.520 EAO with deformable convolution and 0.537$\rightarrow$0.531 EAO with deformable pooling.
	%
	%
	By removing all
	deformable layers, our model can still achieve an EAO of
	0.516, compared favorably against SiamRPN++ with  0.464, suggesting that the proposed 
	Siamese attention and refinement modules are the primary contributors to the performance boost.\\
	
	\begin{table}
		\small
		\setlength{\tabcolsep}{2pt}   	
		\begin{center}
			\begin{tabular}{c|c|ccc}
				\toprule 
				Deformable convolution & Deformable Pooling &  A$\uparrow$ & R$\downarrow$ & EAO$\uparrow$ \\
				\midrule 
				\XSolid & \XSolid       & 0.67 & 0.15 & 0.516 \\
				\XSolid & \checkmark    & 0.67 & 0.15 & 0.520 \\ 
				\checkmark & \XSolid    & 0.68 & 0.15 & 0.531 \\
				\checkmark & \checkmark & 0.68 & 0.14 & 0.537 \\
				\bottomrule	
			\end{tabular}
		\end{center}
		\vspace*{-4mm} 
		\caption{The impact of deformable layers on VOT2016. }
		\vspace{-10pt}
		\label{Tab:deform}
	\end{table}
	
	\noindent\textbf{On Training Data.}
	In this study, we investigate the impact of training  with different training sets. Our current results are achieved by using a combination of multiple training sets from recent LaSOT~\cite{fan2019lasot} and TrackingNet~\cite{muller2018trackingnet}, with COCO~\cite{lin2014microsoft} and YouTube-VOS~\cite{xu2018youtube}, mainly following ~\cite{danelljan2019atom} with an additional YouTube-VOS~\cite{xu2018youtube} for providing mask annotations.    We also report the results on a different  training combination applied by \cite{wang2019fast}, including COCO~\cite{lin2014microsoft}, YouTube-VOS~\cite{xu2018youtube}, YouTube-BoundingBox~\cite{real2017youtube}, ImageNet-VID, and ImageNet-Det~\cite{russakovsky2015imagenet}. Results are reported in Tab.~\ref{Tab:data}. Using the recent large-scale tracking  datasets can improve the results with
	1.2\% EAO on VOT2016, while  our approach can still achieve the state-of-the-art performance using a different choice of the training sets. \\
	
	
	\begin{table}[t]
		\centering
		\resizebox{\columnwidth}{!}{%
			\begin{tabular}{l|l|ccc}
				\toprule
				Method           & Training set                                              & A$\uparrow$   & R$\downarrow$   & EAO$\uparrow$  \\
				\midrule
				SiamAttn         & VID, YTB-BB, COCO, DET, YTB-VOS & 0.68 & 0.15 & 0.525 \\
				SiamAttn         & COCO, YTB
				-VOS, LaSOT, TrackingNet       & 0.68 & 0.14 & 0.537 \\
				\bottomrule
			\end{tabular}
		}
		\caption{Results on VOT2016 with training sets as listed. }
		\label{Tab:data}
		\vspace{-10pt}
	\end{table}
	
	\noindent\textbf{Speed Analysis.}
	On OTB-2015, UAV, LaSOT and TrackingNet benchmarks, our model predicts axis-aligned bounding box, without mask head.
	It can achieve an inference speed of 45 fps.
	On VOT benchmarks, our model generates the rotated boxes from the predicted masks, which reduces the inference speed to 33 fps. 
	
	
	
	\section{Conclusion}\label{sec:cls}
	We have presented new Deformable Siamese Attention Networks for visual object tracking.
	We introduce a deformable Siamese attention mechanism consisting of both self-attention and cross-attention. The new Siamese attention can strongly enhance target discriminability, and at the same time, improve the robustness against large appearance variations, complex backgrounds and distractors. 
	Additionally, a region refinement module is designed to further increase the tracking accuracy. 
	Extensive experiments are conducted on six benchmarks, where our method obtains new state-of-the-art results, with real-time running speed.  
	
	{\small
		\bibliographystyle{ieee_fullname}
		\bibliography{egbib}
	}

	\clearpage
	\setcounter{equation}{0}
\setcounter{figure}{0}
\setcounter{table}{0}
\setcounter{section}{0}

\renewcommand{\theequation}{S\arabic{equation}}
\renewcommand{\thefigure}{S\arabic{figure}}
\renewcommand{\thetable}{S\arabic{table}}
\renewcommand{\thesection}{S\arabic{section}}

\begin{center}
	\textbf{\large Supplementary Material}
\end{center}

	\section{Qualitative Results}
\begin{figure*}[!htb]
		\centering
		\begin{center}
			\includegraphics[width=1\linewidth]{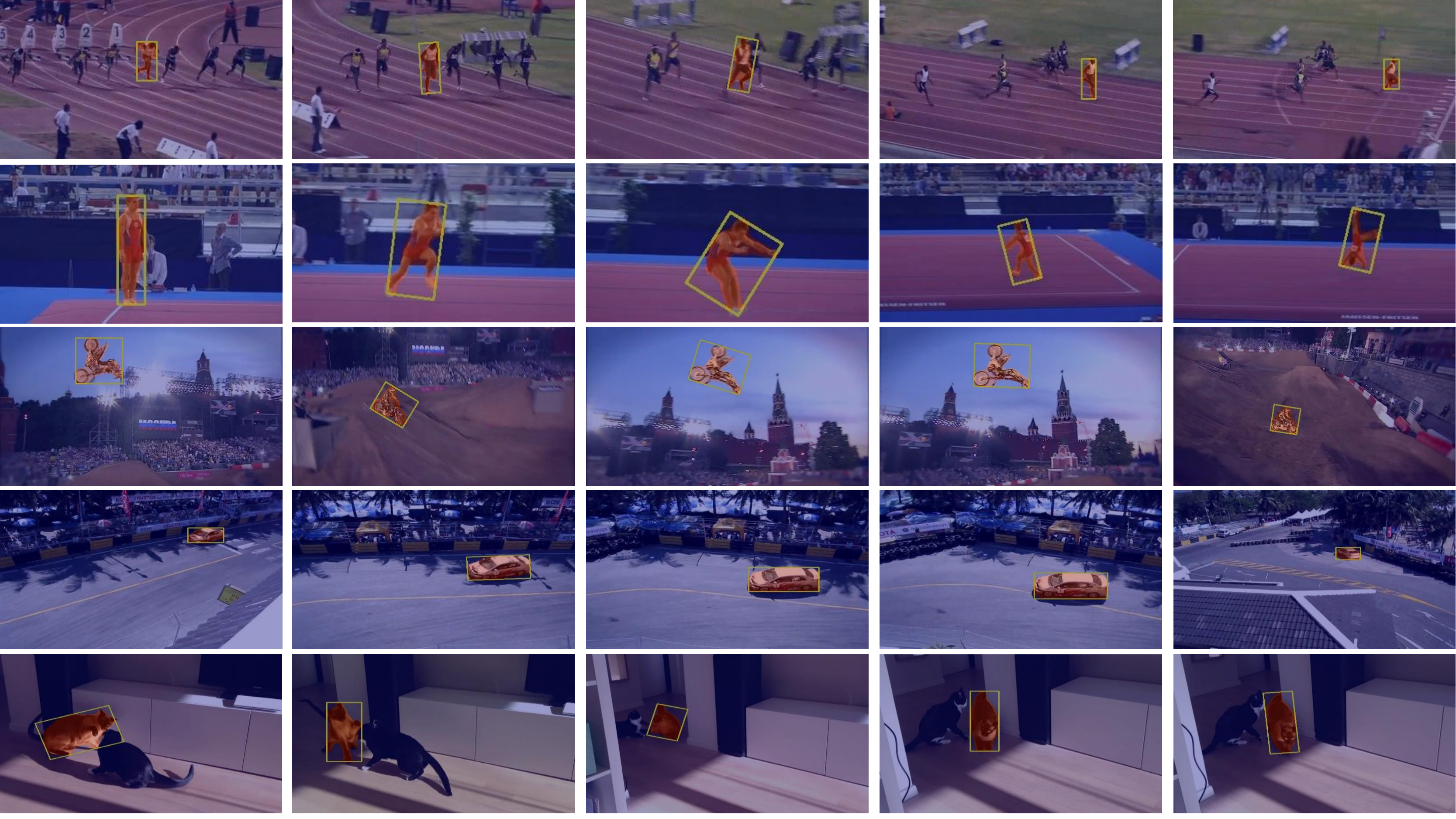}
		\end{center}
		\caption{Qualitative results of SiamAttn on sequences from the visual object tracking benchmark VOT2018.}
		\label{fig:maskexamples}
	\end{figure*}

	Qualitative results of SiamAttn for VOT2018 sequences are shown in Fig.~\ref{fig:maskexamples}. It shows that SiamAttn is capable of tracking and segmenting most objects that have different sizes, motions, deformations and complex background, \etc. 
	
\end{document}